\documentclass[letterpaper, 10 pt, conference]{ieeeconf}  

\IEEEoverridecommandlockouts                              

\usepackage{graphicx} 
\usepackage{epsfig} 
\usepackage{mathptmx} 
\usepackage{times} 
\usepackage{amsmath} 
\usepackage{amssymb}  
\usepackage{hyperref}
\hypersetup{
	pdftitle={No Plan but Everything Under Control: Robustly Solving Sequential Tasks with Dynamically Composed Gradient Descent},
	pdfsubject={We introduce a novel gradient-based approach for solving sequential tasks by dynamically adjusting the underlying myopic potential field in response to feedback and the world's regularities. This adjustment implicitly considers subgoals encoded in these regularities, enabling the solution of long sequential tasks, as demonstrated by solving the traditional planning domain of Blocks World---\emph{without any planning}. Unlike conventional planning methods, our feedback-driven approach adapts to uncertain and dynamic environments, as demonstrated by one hundred real-world trials involving drawer manipulation. These experiments highlight the robustness of our method compared to planning and show how interactive perception and error recovery naturally emerge from gradient descent without explicitly implementing them. This offers a computationally efficient alternative to planning for a variety of sequential tasks, while aligning with observations on biological problem-solving strategies.},
	pdfauthor={Vito Mengers and Oliver Brock},
	pdfkeywords={}
}
\usepackage{cleveref}
\usepackage{siunitx}
\usepackage{todonotes}
\usepackage{algorithm}
\usepackage{algpseudocode}
\usepackage{contour}
\usepackage{contour}

\usepackage{tikz}
\usepackage{setspace}
\usetikzlibrary{decorations.text}
\usetikzlibrary{decorations.markings}
\usetikzlibrary{calc}
\usetikzlibrary{shapes}
\usetikzlibrary{positioning}
\usetikzlibrary{arrows}
\usetikzlibrary{shapes.arrows}
\usetikzlibrary{arrows.meta}
\usetikzlibrary{fit}
\usetikzlibrary{shapes.geometric}
\usetikzlibrary{backgrounds}
\usepackage{braket}
\usepackage{amsmath}
\usepackage{siunitx}
\usepackage{flushend}

\tikzset{
	basic arrow/.style={black, -triangle 60, line width=1pt}
}

\tikzset{
	basic node/.style={draw=black, circle, fill=white, minimum size=30}
}
\tikzset{
	shadow node/.style={draw=white, circle, fill=white, minimum size=30}
}
\tikzset{
	discrete node/.style={draw=black, fill=white, minimum size=30}
}

\tikzset{
	explain/.style={}
}

\definecolor{red}{HTML}{E03131}
\definecolor{blue}{HTML}{4040dd}
\definecolor{yellow}{HTML}{fa8e08}
\definecolor{purple}{HTML}{241623}
\definecolor{green}{HTML}{116a43}
\definecolor{magenta}{HTML}{dc68cf}
\definecolor{cyan}{HTML}{1b998b}
\definecolor{teal}{HTML}{4281a4}
\definecolor{pink}{HTML}{6610f2}
\definecolor{grey}{HTML}{a3b4a2}
\definecolor{changediscrete}{HTML}{44C3A0}
\definecolor{orange}{HTML}{f5b800}
\definecolor{purple}{HTML}{39FF14}
\renewcommand{\contour}[2]{\mathbf{#2}}

\crefname{equation}{eq.}{eqs.}
\crefname{figure}{fig.}{figs.}

\title{\LARGE \bf
	No Plan but Everything Under Control: Robustly Solving Sequential Tasks with Dynamically Composed Gradient Descent
}

\author{Vito Mengers$^{1,2}$ \qquad\ \qquad\ \qquad Oliver Brock$^{1,2}$
	\thanks{$^1$ Robotics and Biology Laboratory, Technische Universit\"at Berlin}
	\thanks{$^2$ Science of Intelligence (SCIoI), Cluster of Excellence, Berlin, Germany}
	\thanks{We gratefully acknowledge funding by the Deutsche Forschungsgemeinschaft (DFG, German Research Foundation) under Germany's Excellence Strategy -- EXC 2002/1 ``Science of Intelligence'' -- project number 390523135.}}%

\Crefname{figure}{Fig.}{Figs.}
\Crefname{section}{Sec.}{Secs.}
\Crefname{equation}{Eq.}{Eqs.}
\DeclareMathOperator*{\argmax}{argmax}

\newcommand{\AICON}{{AICON}}

\usepackage{fancyhdr}

\fancypagestyle{firstpage}{%
	\fancyhead{}
	
	\fancyfoot[C]{{\textcopyright} 2025 IEEE. Published at 2025 IEEE International Conference on Robotics and Automation (ICRA), pp. 90-96.\\
	DOI: \href{https://doi.org/10.1109/ICRA55743.2025.11127552}{10.1109/ICRA55743.2025.11127552}}
}

\begin{document}

\maketitle

\begin{abstract}
We introduce a novel gradient-based approach for solving sequential tasks by dynamically adjusting the underlying myopic potential field in response to feedback and the world's regularities. This adjustment implicitly considers subgoals encoded in these regularities, enabling the solution of long sequential tasks, as demonstrated by solving the traditional planning domain of Blocks World---\emph{without any planning}. Unlike conventional planning methods, our feedback-driven approach adapts to uncertain and dynamic environments, as demonstrated by one hundred real-world trials involving drawer manipulation. These experiments highlight the robustness of our method compared to planning and show how interactive perception and error recovery naturally emerge from gradient descent without explicitly implementing them. This offers a computationally efficient alternative to planning for a variety of sequential tasks, while aligning with observations on biological problem-solving strategies. 
\end{abstract}

\section{Introduction}\thispagestyle{firstpage}

\begin{figure}[t]
	\includegraphics[width=\columnwidth]{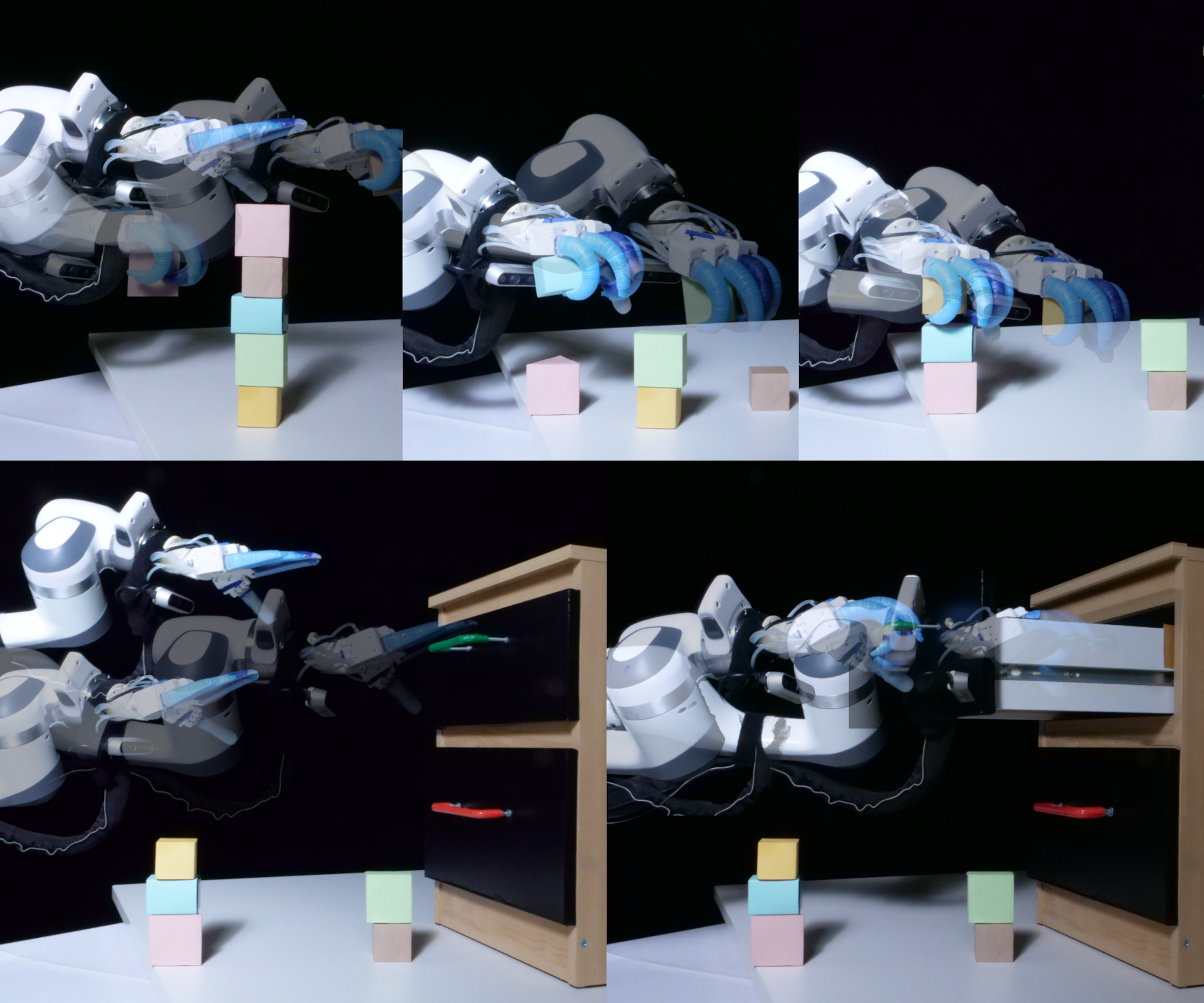}
	\vspace{-4mm}
	\caption{By dynamically composing simpler encoded regularities within gradient descent, our approach aligns actions with both the task and the current state, leading to emergent behavior that needs \emph{no plan to have everything under control}. This allows us to solve long sequential tasks like Blocks World (top) and real-world tasks under uncertainty, such as opening a drawer using only RGB and force-torque measurements (bottom).}\label{fig:title}
	\vspace{-4mm}
\end{figure}

When tasks involve multiple subtasks, sequential steps, or interdependent actions, we often assume that planning is necessary to identify and sequence required subgoals. However, these subgoals and the means to achieve them are inherently embedded in the world's regularities. Therefore, we can solve sequential tasks by relying purely on myopic control if we exploit these regularities effectively. In this paper, we demonstrate how dynamically adjusting the composition of simpler encoded regularities within gradient descent can align actions with the task and the current state of the world, leading to emergent behavior that solves sequential tasks without any planning (\Cref{fig:title}, \href{https://www.tu.berlin/robotics/papers/noplan}{suppl. videos}).

Planning methods also compose regularities~\cite{toussaint_describing_2020}, but they do so by explicitly predicting future states and propagating information through these predictions using search or optimization. This information becomes increasingly uncertain as uncertainty about the current state compounds over longer planning horizons. As a result, costly computed plans often become infeasible, particularly in dynamic or uncertain environments. In contrast, myopic control via gradient descent offers a computationally efficient alternative, adapting quickly to feedback and minimizing uncertainty. However, its application has traditionally been limited to continuous, single-step problems due to the challenge of encoding subgoals into the potential function in a way that accommodates possible variations in task and environment.

Instead of formulating a single potential function to encode all subgoals across all possible variations, we propose to encode the world's regularities that give rise to these subgoals within a set of components and their interconnections. To accommodate task and environmental variations, these interconnections must be active and dynamic, modifying the composition of regularities in response to the perceived state of the world. This dynamic composition ensures that the underlying potential of the gradient continuously evolves, directing the system toward fulfilling the next subgoal that is currently obstructing overall task completion, and allowing necessary subbehaviors to emerge naturally.

Our approach demonstrates several key advantages. By dynamically adjusting interconnections between components, we solve tasks requiring many steps, such as Blocks World, using only myopic gradients (\Cref{sec:blocksworld}). In real-world experiments on drawer manipulation, our method exhibits emergent behaviors like interactive perception and error recovery, maintaining robustness to uncertainty and environmental changes while outperforming traditional planning methods (\Cref{sec:drawer_exp}). Moreover, the distinction between tasks solvable by our approach and those requiring planning aligns with types of biological problem-solving, while our method structurally and behaviorally mirrors biological systems. (\Cref{sec:discussion}).

\section{Related Work}

Robotic behavior generation has traditionally relied on two main approaches: planning-based methods, which predict the future to select action sequences, and gradient-based methods, which adapt actions based on real-time feedback. Here, we review these methods, showing how planning methods struggle with uncertain and dynamic settings, and how we thus aim to extend gradient-based behavior generation to solve tasks that require long sequences of subgoals.

\subsection{Planning-Based Behavior Generation}\label{sec:rw_plan}

Planning has long been a key approach in AI~\cite{fikes_strips_1971,nilsson_principles_1980}, utilizing predictive models based on the world's regularities to select actions that lead to favorable predicted states. Planning in robotics includes explicit search~\cite{jiao_sequential_2022}, planning as inference~\cite{toussaint_probabilistic_2006}, trajectory optimization~\cite{posa_direct_2014}, and sampling-based motion planning~\cite{elbanhawi_sampling-based_2014} as well as its extension into task and motion planning~\cite{kaelbling_integrated_2013,pan_task_2024}. They excel at handling diverse initial states and goals by propagating information on subgoals and interdependencies throughout the trajectory.

However, planning’s reliance on \emph{precise} state estimates and models makes it vulnerable to uncertainty. Over time, compounding uncertainties lead to divergences between plans and real-world execution. Approaches like roadmaps~\cite{payton_internalized_1990,kavraki_probabilistic_1996}, contingency planning~\cite{rhinehart_contingencies_2021}, and belief space planning~\cite{kaelbling_integrated_2013,platt_jr_belief_2010} attempt to mitigate these issues by explicitly considering multiple outcomes or the uncertainty itself during planning. Nevertheless, planning still struggles with unforeseen divergences~\cite{pan_task_2024,garrett_online_2020} and possibly even  detecting critical divergence~\cite{toussaint_dual_2014}, limiting its effectiveness in dynamic real-world settings.

\subsection{Gradient-Based Behavior Generation}\label{sec:rw_gradient}

In contrast, gradient-based behavior generation focuses on myopic control, continuously adapting actions based on real-time feedback without predicting future states. This makes it more suited for uncertain, dynamic environments, as seen in applications like potential fields~\cite{khatib_real-time_1986}, visual servoing~\cite{hutchinson_tutorial_1996}, and model-predictive control~\cite{byravan_se3-pose-nets:_2018}. However, these approaches are prone to local minima, causing robots to become stuck or oscillate~\cite{koren_potential_1991,colotti_determination_2024}. While it is possible to design potential fields without local minima~\cite{e_rimon_exact_1992,connolly_applications_1992}, these are often tailored to specific tasks and environments, reducing their generalizability.

To address this, some methods dynamically reshape the potential~\cite{mabrouk_solving_2008}, set subgoals in it~\cite{bell_forward_2004,lewis_subgoal_1999}, or switch controllers based on the environment~\cite{baum_world_2022}. While effective, they remain task-specific or require predefined subgoals or controllers. Our approach generalizes this by dynamically composing regularities from the world’s structure without predefined subgoals or controllers. Leveraging an actively interconnected perception system~\cite{martin-martin_coupled_2022}, we adapt information flow based on the system’s current state, ensuring that gradient descent leads toward resolving the most relevant subgoal at each step. This allows our system to solve long-horizon, sequential tasks across diverse environments.

\section{{\AICON}: A System for End-to-End Behavior}\label{sec:methods_generic}

Active InterCONnect (\AICON)~\cite{martin-martin_coupled_2022,battaje_information_2024} encodes regularities between sensory inputs ($S$) and actions ($A$) over time ($t$) through \emph{active} interconnections among components, meaning their information exchange changes with the system's state. This enables robust perception, as shown in previous works~\cite{martin-martin_coupled_2022,mengers_combining_2023,pfisterer_helping_2025}. We extend this approach into an end-to-end system that perceives and acts by leveraging these encoded regularities, using gradient descent to pursue goals. As information flow and gradients change, the system dynamically adjusts to resolve the most relevant subgoals. In order to explain this further, we now first describe how we encode regularities in \AICON ~in components (\Cref{sec:components}) and their active interconnections (\Cref{sec:interactions}), before elaborating how we interpret these regularities for action selection with gradient descent in \Cref{sec:gradients}.

\subsection{The Components: Recursive Estimators}\label{sec:components}

Each component estimates a quantity $\mathbf{x}$, representing a part of the world's state, and encodes regularities in ${S \times A \times t}$ to recursively update this estimate. Without any interactions with sensors, actuators, or other components, this remains a poor estimate, which is why the focus of our system lies on \emph{interactions} between components. But before explaining how these interactions are computed in the next section, let us for now assume that they produce a set of informative priors ${\mathbf{c}_1, \mathbf{c}_2, ..., \mathbf{c}_N}$ for the estimated state. With these priors, a component recursively computes its state estimate $\mathbf{x}_t$ based on its previous estimate $\mathbf{x}_{t-1}$ with a differentiable function~$\mathbf{f}$:
\begin{equation}\label{eq:component}
    \mathbf{f}(\mathbf{x}_{t-1}; \mathbf{c}_1, \mathbf{c}_2, ..., \mathbf{c}_N) = \mathbf{x}_{t}~.
\end{equation}

Since each component estimates its state $\mathbf{x}$ with imperfect information, we can also estimate the uncertainty of these estimates. We then treat $\mathbf{f}$ as probabilistic inference, effectively making each component a Bayesian filter encoding regularities in ${S \times A \times t}$. This uncertainty estimation aids in weighting information from various connections and informs action selection, as demonstrated in the drawer task (\Cref{sec:drawer_exp}). However, this probabilistic view is not a requirement of {\AICON} and other forms of recursive estimation are possible, e.g., moving-average filtering.

\subsection{The Interactions: Active Interconnections}\label{sec:interactions}

The quantities estimated by components, sensed by sensors, or executed by actuators are interrelated, forming further regularities in ${S \times A \times t}$. By leveraging these relationships, we enhance robustness of all connected components, because a bidirectional information flow lets each component utilize all available data to refine its estimates. Formally, we can represent each connection between an arbitrary number of quantities ${\mathbf{x}_1, \mathbf{x}_2, ..., \mathbf{x}_M}$ as an implicit differentiable function~$\mathbf{c}$:
\begin{equation}\label{eq:connection}
    \mathbf{c}(\mathbf{x}_1, \mathbf{x}_2, ..., \mathbf{x}_M)~.
\end{equation}

Since $\mathbf{c}$ can depend on multiple quantities, the information exchange between $\mathbf{x}_1$ and $\mathbf{x}_2$ varies with the overall system state and uncertainties. Therefore, we call them \emph{active} and, as we explain in the next subsection, this representation of changing regularities is key to identify subgoals during action selection.

\subsection{The Interpreter for Action Selection: Gradient Descent}\label{sec:gradients}

\AICON ~encodes regularities and extracts information in a way that we can easily use to select goal-directed actions using gradient descent. By defining a goal as a differentiable cost function $g$ of an estimated quantity $\mathbf{x}$, we can compute its gradient $\nabla g(\mathbf{x})$. Using the chain rule, we can then expand it for any quantity in the system through a chain of connections and components, allowing us to determine $\nabla g(\mathbf{a})$ for the actuation signals $\mathbf{a}$. When we descend this gradient employing a gain term $\mathbf{k}$ (\Cref{eq:control}), we can steer the system towards the goal while leveraging the current state estimates and regularities encoded in the system.
\begin{equation}\label{eq:control}
    \mathbf{a}_{t+1} = \mathbf{a}_{t} - \mathbf{k} \cdot \nabla g(\mathbf{a}_t)
\end{equation}

If $\mathbf{c}_{1,2}$ is in a stationary regime, progress towards the goal through it halts. But since we can encode such regime changes in the active interconnection, we can obtain an additional gradient to move into the favorable regime using the chain rule. Formally, if $\mathbf{c}_{1,2}$ also depends on $\mathbf{x}_3$, we derive the gradient $\frac{\partial g}{\partial \mathbf{x}_1} \frac{\partial \mathbf{f}_1}{\partial \mathbf{c}_{1,2}} \frac{\partial \mathbf{c}_{1,2}(\mathbf{x}_1,\mathbf{x}_2,\mathbf{x}_3)}{\partial \mathbf{x}_3}$, thereby replacing the original goal $g$ with a subgoal that seeks to change the regime of $\mathbf{c}_{1,2}$ by adjusting $\mathbf{x}_3$.

This way, we obtain a large set $\mathbb{G}(g,\mathbf{a})$ of gradients from different paths through the network from the goal $g$ to an actuation signal $\mathbf{a}$. They represent different direct ways of reaching the goal as well as different subgoals of steering the system into more favorable regimes. Since they represent often mutually exclusive ways of acting, we cannot simply use their sum for gradient descent as we would often fall victim to the problem of conflicting gradients~\cite{liu_conflict-averse_2021}. Hence, instead of trying to find an optimal mixture, we only select the currently steepest gradient at any given time (\Cref{eq:steepest}). This always selects an appropriate action, because the gradient magnitude is scaled according to the goal $g$ and system's state while gradients stuck at stationary points have by definition near-zero magnitude.
\begin{equation}\label{eq:steepest}
	\nabla^* g(\mathbf{a}_t) = \argmax_{\nabla g(\mathbf{a}_t) \in \mathbb{G}(g,\mathbf{a}_t)} \|\nabla g(\mathbf{a}_t)\|~.
\end{equation}

In summary, at each time step, we update system estimates (\Cref{eq:component,eq:connection}), and use the steepest gradient to adjust actions (\Cref{eq:control,eq:steepest}). This way, we leverage encoded regularities for robust state estimation and action selection in an integrated manner. Furthermore, it allows us to swiftly adapt both estimation and action selection based on new environmental feedback and to dynamically set subgoals for changing the current regularity regime. In the following sections, we first demonstrate how \AICON ~solves long sequential tasks in the classic symbolic AI domain of Blocks World and then explore its application to uncertain and dynamic settings through a real-world partially observable drawer-opening task.

\begin{figure*}[t]
	\newcommand{\myfrac}[2]{{\footnotesize\genfrac{}{}{}{}{\raisebox{1pt}{$#1$}}{\raisebox{-2pt}{$#2$}}}}
	\newcommand{\sidedist}{0.8cm}
	\newcommand{\goaldist}{0.05cm}
	\newcommand{\goalsidedist}{1.1cm}
	\newcommand{\seplinedist}{1.125cm}
	\newcommand{\nodedist}{0cm}
	\newcommand{\nodesize}{0.3cm}
	\newcommand{\topdist}{-0.1cm}
	\newcommand{\goalmathshift}{0cm}
	\newcommand{\arrowshift}{0.1cm}
	\newcommand{\framesep}{0.005cm}
	\begin{minipage}{0.6821\linewidth}
		\begin{tikzpicture}[node distance=\nodedist,minimum size=\nodesize,draw=white,ultra thin,font=\footnotesize,rounded corners=2pt,framed,inner frame sep=\framesep]
			
			\node [rectangle,fill=yellow] (dgoal) {};
			\node [rectangle,above=0cm of dgoal,fill=red] (agoal) {};
			\node [rectangle,above=0cm of agoal,fill=white] (helper) {};
			
			\node [above left=-0.02cm and 0.24cm of helper,minimum size=0.41cm] (mark) {\small\textbf{a}};
			
			\node [below=\goaldist of dgoal, align=right,xshift=\goalmathshift] () {$g=o(\contour{black}{\textcolor{red}{R}},\contour{black}{\textcolor{yellow}{O}})$};
			
			\node [right=\seplinedist of mark,minimum size=0cm,yshift=0cm] (mark) {};
			\node [below=1.62cm of mark,minimum size=0cm] (mark2) {};
			\draw[dashed,draw=gray,thick] (mark.north) -- (mark2.south);
			
			\node [rectangle,fill=red,right=\goalsidedist of dgoal] (a) {};
			\node [rectangle,above=0cm of a,fill=green] (b) {};
			\node [rectangle,above=0cm of b,fill=blue] (c) {};
			
			\node [rectangle,fill=yellow, right=of a] (d) {};
			\node [rectangle,above=0cm of d,fill=cyan] (e) {};
			
			\node [rectangle,fill=orange,right=of d] (f) {};
			\node [rectangle,above=0cm of f,fill=magenta] (g) {};
			\node [rectangle,above=0cm of g,fill=teal] (h) {};
			
			\draw [-latex,draw=black,out=180,in=70] (c.west) to ([xshift=-0.2cm, yshift=-0.2cm]c.west);
			
			\node [rectangle,above=0cm of e,fill=white] (helper) {};
			\node [above=0cm of helper,minimum size=0cm,align=left,yshift=-0.1cm] () {\scriptsize \textit{Subgoal} $c(\contour{black}{\textcolor{red}{R}})$};

			\node [below=\topdist of a] (s1) {};
			
			\node [rectangle,fill=red,right=\sidedist of f] (a) {};
			\node [rectangle,above=0cm of a,fill=green] (b) {};
			
			\node [rectangle,fill=yellow, right=of a] (d) {};
			\node [rectangle,above=0cm of d,fill=cyan] (e) {};
			
			\node [rectangle,fill=orange,right=of d] (f) {};
			\node [rectangle,above=0cm of f,fill=magenta] (g) {};
			\node [rectangle,above=0cm of g,fill=teal] (h) {};
			
			\node [rectangle,right=of f,fill=blue] (c) {};
			
			\draw [-latex,draw=black,out=180,in=70] (b.west) to ([xshift=-0.2cm, yshift=-0.2cm]b.west);
			
			\node [above=0cm of h,minimum size=0cm,align=left,yshift=-0.1cm] () {\scriptsize \textit{Subgoal} $c(\contour{black}{\textcolor{red}{R}})$};

			\node [below=\topdist of d] (s2) {};
			\draw [-latex,draw=black] ([xshift=\arrowshift]s1.west) to node [below] {$\myfrac{\partial o(\contour{black}{\textcolor{red}{R}},\contour{black}{\textcolor{yellow}{O}})}{\partial c(\contour{black}{\textcolor{red}{R}})}\myfrac{\partial c(\contour{black}{\textcolor{red}{R}})}{\partial o(\contour{black}{\textcolor{blue}{B}},\contour{black}{\textcolor{red}{R}})}$} (s2.west);
			
			\node [rectangle,right=\sidedist of c,fill=green] (b) {};
			\node [rectangle,fill=red,right=of b] (a) {};
			
			\node [rectangle,fill=yellow, right=of a] (d) {};
			\node [rectangle,above=0cm of d,fill=cyan] (e) {};
			
			\node [rectangle,fill=orange,right=of d] (f) {};
			\node [rectangle,above=0cm of f,fill=magenta] (g) {};
			\node [rectangle,above=0cm of g,fill=teal] (h) {};
			
			\node [rectangle,right=of f,fill=blue] (c) {};
			
			\draw [-latex,draw=black,out=90,in=-20] (e.north) to ([xshift=-0.2cm, yshift=0.2cm]e.north);
			
			\node [above=0cm of h,minimum size=0cm,align=left,yshift=-0.1cm] () {\scriptsize \textit{Subgoal}  $c(\contour{black}{\textcolor{yellow}{O}})$};

			\node [below=\topdist of f] (s3) {};
			\draw [-latex,draw=black] ([xshift=\arrowshift]s2.west) to node [below] {$\myfrac{\partial o(\contour{black}{\textcolor{red}{R}},\contour{black}{\textcolor{yellow}{O}})}{\partial c(\contour{black}{\textcolor{red}{R}})}\myfrac{\partial c(\contour{black}{\textcolor{red}{R}})}{\partial o(\contour{black}{\textcolor{green}{G}},\contour{black}{\textcolor{red}{R}})}$} (s3.east);
			
			\node [rectangle,right=\sidedist of c,fill=green] (b) {};
			\node [rectangle,fill=red,right=of b] (a) {};
			
			\node [rectangle,fill=yellow, right=of a] (d) {};
			
			\node [rectangle,fill=orange,right=of d] (f) {};
			\node [rectangle,above=0cm of f,fill=magenta] (g) {};
			\node [rectangle,above=0cm of g,fill=teal] (h) {};
			
			\node [rectangle,right=of f,fill=blue] (c) {};
			\node [rectangle,right=of c,fill=cyan] (e) {};
			
			\draw [-latex,draw=black,out=90,in=110] (a.north) to (d.north);

			\node [below=\topdist of f] (s4) {};
			\draw [-latex,draw=black] ([xshift=\arrowshift]s3.east) to node [below] {$\myfrac{\partial o(\contour{black}{\textcolor{red}{R}},\contour{black}{\textcolor{yellow}{O}})}{\partial c(\contour{black}{\textcolor{yellow}{O}})}\myfrac{\partial c(\contour{black}{\textcolor{yellow}{O}})}{\partial o(\contour{black}{\textcolor{cyan}{C}},\contour{black}{\textcolor{yellow}{O}})}$} (s4.center);
			
			\node [rectangle,right=\sidedist of e,fill=green] (b) {};
			\node [rectangle,fill=yellow, right=of b] (d) {};
			\node [rectangle,fill=red, above=0cm of d] (a) {};

			\node [rectangle,fill=orange,right=of d] (f) {};
			\node [rectangle,above=0cm of f,fill=magenta] (g) {};
			\node [rectangle,above=0cm of g,fill=teal] (h) {};
			
			\node [rectangle,right=of f,fill=blue] (c) {};
			\node [rectangle,right=of c,fill=cyan] (e) {};

			\node [below=\topdist of f] (s5) {};
			\draw [-latex,draw=black] ([xshift=\arrowshift]s4.center) to node [below] {$\myfrac{\partial g}{\partial o(\contour{black}{\textcolor{red}{R}},\contour{black}{\textcolor{yellow}{O}})}$} (s5.center);
			
			\node [below=0cm of s5,minimum size=0cm,align=left,yshift=0.12cm,xshift=0.22cm] () {\scriptsize \textit{Goal} $g$\\\scriptsize\textit{Reached}};
		\end{tikzpicture}
		
		\begin{tikzpicture}[node distance=\nodedist,minimum size=\nodesize,draw=white,ultra thin,font=\footnotesize,rounded corners=2pt,framed,inner frame sep=\framesep]
			
			\node [rectangle,fill=red] (agoal) {};
			\node [rectangle,above=0cm of agoal,fill=green] (bgoal) {};
			\node [rectangle,above=0cm of bgoal,fill=blue] (cgoal) {};
			\node [left=0.24cm of cgoal,minimum size=0.41cm] (mark) {\small\textbf{b}};

			\node [below=\goaldist of agoal,align=right,xshift=\goalmathshift] () {$g=o(\contour{black}{\textcolor{green}{G}},\contour{black}{\textcolor{red}{R}})$\\$+o(\contour{black}{\textcolor{blue}{B}},\contour{black}{\textcolor{green}{G}})$};
			
			\node [right=\seplinedist of mark,minimum size=0cm,yshift=0.05cm] (mark) {};
			\node [below=1.3cm of mark,minimum size=0cm] (mark2) {};
			\draw[dashed,draw=gray,thick] (mark.north) -- (mark2.south);
			
			\node [rectangle,fill=red,right=\goalsidedist of agoal] (a) {};
			\node [rectangle,right=of a,fill=green] (b) {};
			\node [rectangle,right=of b,fill=blue] (c) {};
			
			\draw [-latex,draw=black,out=90,in=70] (c.north) to (b.north);
			
			\node [below=\topdist of a] (s1) {};
			
			\node [right=\sidedist of c] (empty) {};
			\node [rectangle,fill=red,right=of empty] (a) {};
			\node [rectangle,right=of a,fill=green] (b) {};
			\node [rectangle,above=0cm of b,fill=blue] (c) {};
			\node [right=of b] (empty) {};
			
			\draw [-latex,draw=black,out=90,in=200] (c.north) to ([xshift=0.2cm, yshift=0.2cm]c.north);
			
			\node [above right=-0.3cm and -0.0cm of c,minimum size=0cm,align=center,yshift=-0.12cm] () {\scriptsize \textit{Subgoal}\\\scriptsize $c(\contour{black}{\textcolor{green}{G}})$};
			
			\node [below=\topdist of a] (s2) {};
			\draw [-latex,draw=black] ([xshift=\arrowshift]s1.west) to node [below] {$\myfrac{\partial g}{\partial o(\contour{black}{\textcolor{blue}{B}},\contour{black}{\textcolor{green}{G}})}$} (s2.west);
			
			\node [right=\sidedist of empty] (empty) {};
			\node [rectangle,fill=red,right=of empty] (a) {};
			\node [rectangle,right=of a,fill=green] (b) {};
			\node [rectangle,right=of b,fill=blue] (c) {};
			\node [right=of c] (empty) {};
			
			\draw [-latex,draw=black,out=90,in=70] (b.north) to (a.north);

			\node [below=\topdist of c] (s3) {};
			\draw [-latex,draw=black] ([xshift=\arrowshift]s2.west) to node [below] {$\myfrac{\partial g}{\partial o(\contour{black}{\textcolor{green}{G}},\contour{black}{\textcolor{red}{R}})}\myfrac{\partial c(\contour{black}{\textcolor{green}{G}})}{\partial o(\contour{black}{\textcolor{blue}{B}},\contour{black}{\textcolor{green}{G}})}$} (s3.east);
			
			\node [right=\sidedist of empty] (empty) {};
			\node [right=of empty] (empty) {};
			\node [rectangle,fill=red,right=of empty] (a) {};
			\node [rectangle,above=0cm of a,fill=green] (b) {};
			\node [rectangle,right=of a,fill=blue] (c) {};
			\node [right=of c] (empty) {};
			\node [right=of empty] (empty) {};
			
			\draw [-latex,draw=black,out=90,in=70] ([yshift=0.3cm]c.north)
			to (b.north);
			
			\draw [draw=black] (c.north) to ([yshift=0.3cm]c.north);

			\node [below=\topdist of c] (s4) {};
			\draw [-latex,draw=black] ([xshift=\arrowshift]s3.east) to node [below] {$\myfrac{\partial g}{\partial o(\contour{black}{\textcolor{green}{G}},\contour{black}{\textcolor{red}{R}})}$} (s4.center);
			
			\node [right=\sidedist of empty] (empty) {};
			\node [right=of empty] (empty) {};
			\node [rectangle,fill=red,right=of empty] (a) {};
			\node [rectangle,above=0cm of a,fill=green] (b) {};
			\node [rectangle,above=0cm of b,fill=blue] (c) {};

			\node [below=\topdist of a] (s5) {};
			\draw [-latex,draw=black] ([xshift=\arrowshift]s4.center) to node [below] {$\myfrac{\partial g}{\partial o(\contour{black}{\textcolor{blue}{B}},\contour{black}{\textcolor{green}{G}})}$} (s5.center);
			
			\node [below=0cm of s5,minimum size=0cm,align=left,yshift=-0.05cm,xshift=0.22cm,text height=0.33cm] () {\scriptsize \textit{Goal} $g$\\\scriptsize\textit{Reached}};
			
		\end{tikzpicture}
	\end{minipage}
	\begin{minipage}{0.2865\linewidth}
		\begin{tikzpicture}[node distance=\nodedist,minimum size=\nodesize,draw=white,ultra thin,font=\footnotesize,rounded corners=2pt,framed,inner frame sep=\framesep]
			\node[minimum size=0cm] (img) {\includegraphics[width=\linewidth]{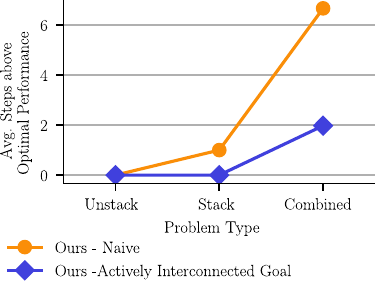}};
			\node [minimum size=0cm,above left=-0.5cm and -0.25cm of img] () {\normalsize\textbf{c}};

		\end{tikzpicture}
	\end{minipage}
	\caption{Dynamically composed gradient descent solves the classic planning domain of \emph{Blocks World} (BW): 
	(\textbf{a}) Gradient descent with an \AICON-ic system always selects actions that advance towards the goal, identifying necessary subgoals while ignoring irrelevant blocks.
	(\textbf{b}) But a naive goal formulation can lead to suboptimal actions and necessary backtracking due to competing subgoals.
	(\textbf{c}) Actively interconnecting the goal with the current state removes competing subgoals, though some efficient shortcuts in combined problems remain challenging to find without foresight. On a set of $130$ randomly generated BW problems, we show how our approach consistently finds solutions and closely approaches optimal performance without direct state predictions.} \label{fig:blocksworld}
	\vspace*{-4mm}
\end{figure*}
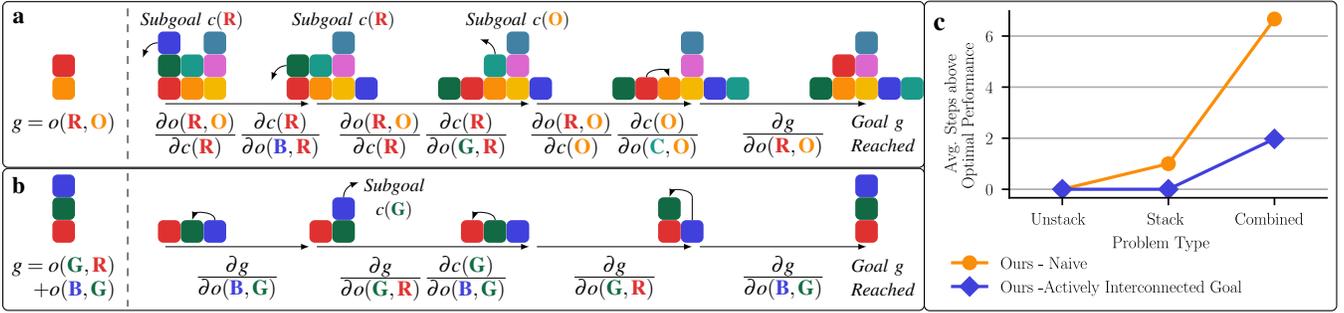

\section{An \AICON-ic System for Blocks World}\label{sec:blocksworld}

To investigate the kinds of problems solvable by dynamically composed gradient descent, we turn to an (in)famous planning domain: \emph{Blocks World} (BW)~\cite{nilsson_principles_1980,slaney_blocks_2001}, where blocks must be moved one at a time to form a goal state from an initial arrangement (see \Cref{fig:blocksworld}). 

While many planners have tackled BW, to our knowledge, gradient descent has not been applied, likely due to the need for numerous subgoals. However, since BW subgoals arise from block relationships, \AICON's ability to dynamically react to these relationships offers a novel solution. In the following, we describe the simple \AICON-ic system for BW (\Cref{sec:bw_network}), explain the resulting gradients (\Cref{sec:bw_grads}), and evaluate the system on various BW instances (\Cref{sec:bw_exp}).

\subsection{Components and Active Interconnections}\label{sec:bw_network}
To model BW, we follow standard approaches~\cite{slaney_blocks_2001}, considering two states: a block $X$ can be \emph{on} another block $Y$, and \emph{clear} if no block is on top of it. To ensure differentiability, we define likelihoods $o(X,Y)$ and $c(X)$ for these states, updating them recursively based on actions $a_\mathrm{stack}$ and $a_\mathrm{unstack}$, as shown in \Cref{eq:bw_clear,eq:bw_on}, where $\mathbb{B}$ is the set of all blocks.
\begin{gather}
	c_t(X) = 1 - \frac{1}{\|\mathbb{B}\|}\sum_{Y \in \mathbb{B}} o_t(Y,X)	\label{eq:bw_clear}\\
	\begin{split}
	o_t(X,Y) = o_{t-1}(X,Y)
		&+ c_{t-1}(X) \cdot c_{t-1}(Y) \cdot a_\mathrm{stack}(X,Y)\\
		&- c_{t-1}(X) \cdot a_\mathrm{unstack}(X,Y)
	\end{split} \label{eq:bw_on}
\end{gather}

Here, $c(X)$ depends on $o(X,Y)$ across all blocks, while $o(X,Y)$ is influenced by $c(X)$ and $c(Y)$, creating active interconnections that regulate how actions affect the current state.

\subsection{Resulting Gradients and Behavior}\label{sec:bw_grads}
To solve a BW problem using \AICON, we define a goal through a differentiable cost function $g$ based on the states of the blocks. Gradients derived from this cost function guide the system's actions. In BW, these gradients generally fall into two categories: $\nabla_\mathrm{stack}$ (\Cref{eq:bw_stack}), which directs the system to stack one block on another as part of the goal, and $\nabla_\mathrm{unstack}$ (\Cref{eq:bw_unstack}), which directs the system to remove a block to clear an underlying one.
\begin{align}
	&\nabla_\mathrm{stack}=\frac{\partial g}{\partial o(X,Y)}\frac{\partial o(X,Y)}{\partial a_\mathrm{stack}(X,Y)}\label{eq:bw_stack}\\
	&\nabla_\mathrm{unstack}=\frac{\partial g}{\partial o(X,Y)}\frac{\partial o(X,Y)}{\partial c(X)}\frac{\partial c(X)}{\partial o(Z,X)}\frac{\partial o(Z,X)}{\partial a_\mathrm{unstack}(Z,X)}\label{eq:bw_unstack}
\end{align}

Depending on the world state and goal, these gradients will either be $0$ or $1$ for different block combinations, representing feasible actions toward the goal. This allows solving complex BW tasks efficiently without unnecessary moves (\Cref{fig:blocksworld}a). However, in the presence of multiple feasible actions, the system randomly selects, sometimes resulting in suboptimal outcomes when competing subgoals arise (\Cref{fig:blocksworld}b). To mitigate this, we can refine goals to minimize conflicts, for example by prioritizing actions based on the lowest unstacked block, creating an additional layer of interconnection between the goal and states.  We will further investigate these strategies and their impact on performance in the subsequent section.

\subsection{Performance on Various Blocks World Instances}\label{sec:bw_exp}

We generated over $100$ BW instances with 10 to 30 blocks involving 0 to 15 towers, following~\cite{slaney_blocks_2001}, and compared our basic gradient descent and its variant with actively interconnected goals to optimal solutions. Remarkably, both methods solved all instances, including those requiring at least 35 steps, and nearly achieved optimal performance without any predictions.

As seen in \Cref{fig:blocksworld}c, gradient descent consistently solves unstacking tasks optimally, since they lack competing subgoals. However, for stacking tasks, naive goal formulations lead to suboptimality due to competing subgoals. Although the interconnected goal version improves performance, it still struggles with identifying useful shortcuts like bypassing intermediate steps by moving blocks directly between towers. Consistently finding these shortcuts requires planning, but is recognized as an NP-hard planning problem~\cite{chenoweth_np-hardness_1991}.

In summary, our approach transforms planning problems into gradient descent tasks by dynamically adapting potentials to the current state. While this efficiently identifies subgoals, it struggles with competing ones, leading to some suboptimalities. Instead of guaranteeing optimality, we can, however, easily adapt to feedback, which allows robust performance in real-world tasks, as we show in the next section.

\begin{figure*}[t]
	\includegraphics[width=\textwidth]{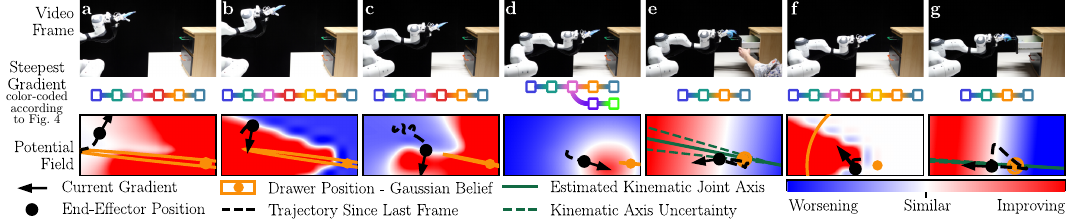}
	\vspace{-6mm}
	\caption{The dynamically composed gradient descent adapts the potential field depending on the current situation: (\textbf{a}) With initial uncertainty about the drawer’s position, the robot adjusts its viewpoint to reduce uncertainty. (\textbf{b}) If it looses sight of the drawer, it returns to the visible area. (\textbf{c}) This continues to further reduce uncertainty, continuously adapting the potential to guide to novel viewpoints. (\textbf{d}) Once the drawer's position is sufficiently certain, the robot approaches and then grasps it. (\textbf{e}) The robot then moves the drawer along its estimated kinematic axis, but we disturb the opening causing it to loose the drawer. (\textbf{f}) This increases uncertainty, necessitating visual re-identification. (\textbf{g}) After ensuring the drawer's position anew, the robot can then grasp and open the drawer along kinematic joint axis while improving its estimate of the axis parameters. We visualize the potential field by restricting the robot's motion to the xz-plane and spatially sampling the gradient across the plane post-execution. The steepest gradient is color-coded according to \Cref{fig:network}. For a video version of both potential and the current gradient path through the network, see the \href{https://www.tu.berlin/robotics/papers/noplan}{suppl. videos}.}\label{fig:potentials}
	\vspace{-4mm}
\end{figure*}

\begin{figure}[b!]
	\vspace{-3mm}
	\newcommand{\myfrac}[2]{{\footnotesize\genfrac{}{}{}{}{\raisebox{1pt}{$#1$}}{\raisebox{-2pt}{$#2$}}}}
	\newcommand{\sidedist}{0.8cm}
	\newcommand{\goaldist}{0.05cm}
	\newcommand{\goalsidedist}{1.1cm}
	\newcommand{\seplinedist}{1.125cm}
	\newcommand{\nodedist}{0.23cm}
	\newcommand{\nodesize}{1.4cm}
	\newcommand{\topdist}{-0.1cm}
	\newcommand{\goalmathshift}{0cm}
	\newcommand{\arrowshift}{0.1cm}
	\newcommand{\framesep}{0.005cm}
	\tikzset{test/.style n args={3}{
			postaction={
				decorate,
				decoration={
					markings,
					mark=between positions 0 and \pgfdecoratedpathlength step 0.5pt with {
						\pgfmathsetmacro\myval{multiply(
							divide(
							\pgfkeysvalueof{/pgf/decoration/mark info/distance from start}, \pgfdecoratedpathlength
							),
							100
							)};
						\pgfsetfillcolor{#3!\myval!#2};
						\pgfpathcircle{\pgfpointorigin}{#1};
						\pgfusepath{fill};}
	}}}}
	\begin{tikzpicture}[node distance=\nodedist,minimum width=\nodesize,minimum height=0.55cm,draw=black, thin,font=\small,rounded corners=2pt]
		\node [rectangle, draw=teal, ultra thick] (aee) {$\mathbf{a}_\mathrm{ee}$};
		\node [rectangle, draw=black, below=of aee] (zee) {$\mathbf{z}_\mathrm{ee}$};
		\node [rectangle, fill=gray!20, decoration={
			markings,
			mark=at position 0.2 with {\arrow{Triangle}},
			mark=at position 0.7 with {\arrow{Triangle}}
		}, draw=black, thick,
		postaction={decorate}, minimum height=1.56cm, minimum width=1.6cm, right=of aee, yshift=-0.4cm, xshift=-0.1cm] (eerec) {};
		\node [rectangle, draw=yellow, right=of aee,ultra thick,fill=white] (xee) {$\mathbf{x}_\mathrm{ee}$};
		\node [rectangle, draw=black, below=of xee, fill=white] (sigmaee) {$\boldsymbol{\Sigma}_\mathrm{ee}$};
		
		\node [rectangle, draw=black, above=of aee] (zft) {$\mathbf{z}_\mathrm{ft}$};

		\node [rectangle, draw=purple, above=of zft, ultra thick] (ahand) {$a_\mathrm{hand}$};
		\node [rectangle, draw=black, above=of ahand] (zhand) {$z_\mathrm{hand}$};
		\node [rectangle, fill=gray!20, decoration={
			markings,
			mark=at position 0.2 with {\arrow{Triangle}},
			mark=at position 0.7 with {\arrow{Triangle}}
		}, draw=black, thick,
		postaction={decorate}, minimum height=1.56cm, minimum width=1.6cm, right=of zhand, yshift=-0.4cm, xshift=-0.1cm, ] (handrec) {};
		\node [rectangle, draw=pink, right=of ahand,fill=white, ultra thick] (xhand) {$x_\mathrm{hand}$};
		\node [rectangle, draw=black, above=of xhand,fill=white] (sigmahand) {${\sigma}_\mathrm{hand}$};
		
		\node [rectangle, draw=black, below=of zee] (zrgb) {$\mathbf{z}_\mathrm{rgb}$};
		\node [rectangle, draw=white, right=of zrgb] (helpvis) {};
		\node [rectangle, draw=white, right=of zft] (helpgrasp) {};
		
		\node [rectangle, fill=gray!20, decoration={
			markings,
			mark=at position 0.27 with {\arrow{Triangle}},
			mark=at position 0.77 with {\arrow{Triangle}}
		}, draw=black, thick,
		postaction={decorate}, minimum height=0.8cm, minimum width=1.6cm, right=of helpgrasp, yshift=-0cm, xshift=-0.05cm] (grasprec) {};
		\node [rectangle, draw=magenta, right=of helpgrasp, ultra thick,fill=white] (grasp) {$p_\mathrm{grasped}$};
		\node [rectangle, fill=gray!20, decoration={
			markings,
			mark=at position 0.27 with {\arrow{Triangle}},
			mark=at position 0.77 with {\arrow{Triangle}}
		}, draw=black, thick,
		postaction={decorate}, minimum height=0.8cm, minimum width=1.6cm, right=of helpvis, yshift=-0cm, xshift=-0.05cm] (visrec) {};
		\node [rectangle, draw=orange, right=of helpvis, ultra thick,fill=white,xshift=0.05cm] (visible) {$p_\mathrm{visible}$};

		\node [rectangle, draw=white, right=of xee] (drawerhelp) {};
		\node [rectangle, fill=gray!20, decoration={
			markings,
			mark=at position 0.2 with {\arrow{Triangle}},
			mark=at position 0.7 with {\arrow{Triangle}}
		}, draw=black, thick,
		postaction={decorate}, minimum height=1.56cm, minimum width=1.6cm, right=of drawerhelp, yshift=-0.4cm, xshift=-0.1cm] (drawerrec) {};
		\node [rectangle, draw=green, right=of drawerhelp,ultra thick, fill=white] (xdrawer) {$\mathbf{x}_\mathrm{drawer}$};
		\node [rectangle, draw=red, below=of xdrawer,ultra thick, fill=white] (sigmadrawer) {$\boldsymbol{\Sigma}_\mathrm{drawer}$};
		
		\node [rectangle, draw=white, right=of xhand] (jointhelp) {};
		\node [rectangle, draw=white, above=of jointhelp] (jointrechelp) {};
		\node [rectangle, minimum height=1.56cm, minimum width=1.6cm, right=of jointrechelp, yshift=-0.4cm, xshift=-0.1cm, fill=gray!20, decoration={
			markings,
			mark=at position 0.2 with {\arrow{Triangle}},
			mark=at position 0.7 with {\arrow{Triangle}}
		}, draw=black, thick,
		postaction={decorate}] (jointrec) {};
		\node [rectangle, draw=cyan, right=of jointhelp,ultra thick, fill=white] (xjoint) {$\mathbf{x}_\mathrm{joint}$};
		\node [rectangle, draw=black, above=of xjoint,fill=white] {$\boldsymbol{\Sigma}_\mathrm{joint}$};
		\node [ellipse, draw=blue, right=of xjoint,yshift=0.44cm,minimum width=0cm,minimum height=0cm, ultra thick] (g) {Goal $g$};

		\node [circle, draw=black,minimum width=0cm,minimum height=0.3cm, ultra thick] (ai1) at ($(visible)!0.5!(grasp)$) {\small$\mathbf{\sim}$};
		\node [circle, draw=black,minimum width=0cm,minimum height=0.3cm, ultra thick] (ai2) at ($(xjoint)!0.5!(xdrawer)$) {\small$\sim$};
		\node [circle, draw=black,minimum width=0cm,minimum height=0.3cm, ultra thick] (ai3) at ($(xee)!0.5!(xhand)$) {\small$\sim$};
		
		\draw [ultra thick,black] (eerec) to (ai1);
		\draw [ultra thick,black] (drawerrec) to (ai1);
		\draw [ultra thick,black] (grasprec) to (ai1);
		\draw [ultra thick,black] (visrec) to (ai1);
		\draw [ultra thick,black] (zrgb) to ([xshift=-0.25cm]helpvis.east) to (ai1);

		\draw [ultra thick,black] (jointrec) to (ai2);
		\draw [ultra thick,black] (drawerrec) to (ai2);
		\draw [ultra thick,black] (grasprec) to (ai2);

		\draw [ultra thick,black] (grasprec) to (ai3);
		\draw [ultra thick,black] (zft) to (ai3);
		\draw [ultra thick,black] (handrec) to (ai3);
		
		\draw [ultra thick,black] (zhand) to (handrec);
		\draw [ultra thick,black] (ahand) to (handrec);
		
		\draw [ultra thick,black] (zee) to (eerec);
		\draw [ultra thick,black] (aee) to (eerec);
		
		\draw [test={1.6pt}{blue}{cyan}, thick] (g) to (xjoint);
		\draw [test={1.6pt}{cyan}{green}, thick] ([xshift=-0.1cm]xjoint.south) to ([xshift=-0.1cm]xdrawer.north);
		\draw [test={1.6pt}{green}{yellow}, thick] (xdrawer) to (xee);
		\draw [test={1.6pt}{cyan}{magenta}, thick] ([xshift=-0.1cm]xjoint.south) to [bend left=30] (grasp.east);
		\draw [test={1.6pt}{magenta}{pink}, thick] (grasp.west) to [bend left=30] ([xshift=0.1cm]xhand.south);
		\draw [test={1.6pt}{pink}{purple}, thick] (xhand) to  (ahand);
		\draw [test={1.6pt}{magenta}{yellow}, thick] ([xshift=-0.1cm]grasp.south) to [bend left=30] (xee.east);
		\draw [test={1.6pt}{magenta}{red}, thick] ([xshift=0.1cm]grasp.south) to [bend right] (sigmadrawer.west);
		\draw [test={1.6pt}{red}{orange}, thick] (sigmadrawer.west) to [bend right=30] ([xshift=0.1cm]visible.north);
		\draw [test={1.6pt}{red}{yellow}, thick] (sigmadrawer.west) to (xee.east);
		\draw [test={1.6pt}{orange}{yellow}, thick] ([xshift=-0.1cm]visible.north) to [bend right=30] (xee.east);
		\draw [test={1.6pt}{yellow}{teal}, thick] (xee) to (aee);

		\node [right=0.2cm of xdrawer, rectangle, fill=gray!20, decoration={
			markings,
			mark=at position 0.2 with {\arrow{Triangle}},
			mark=at position 0.7 with {\arrow{Triangle}}
		}, draw=black,
		postaction={decorate}, minimum width=0.5cm, minimum height=0.3cm,yshift=0.4cm] (lrec) {};
		\node [right=0.02cm of lrec, align=left] {\scriptsize Recursive\\[-0.9ex]\scriptsize Estimator~~~~~~~~~~~~~~~};
		\node [above=0.4cm of lrec, rectangle, fill=white, minimum width=0.5cm, minimum height=0.3cm, draw=black] (lquant) {};
		\node [right=0.02cm of lquant, align=left] {\scriptsize Quantity~~~~~~~};
		\node [below=0.5cm of lrec, rectangle, minimum width=0.5cm, minimum height=0.1cm, draw=blue, thick] (lgrad1) {};
		\node [below=0.1cm of lgrad1, rectangle, minimum width=0.5cm, minimum height=0.1cm, draw=orange, thick] (lgrad2) {};
		\draw [test={0.8}{blue}{orange}, thick] (lgrad1) to (lgrad2);
		\node [right=0.02cm of lgrad1, align=left,yshift=-0.2cm] {\scriptsize Gradient\\[-0.9ex]\scriptsize between\\[-0.9ex]\scriptsize \textcolor{blue}{Blue} and\\[-0.9ex]\scriptsize \textcolor{orange}{Yellow}\\[-0.9ex]\scriptsize Quantity~~~~~~~~~~~~~~~};
		\node [below=0.5cm of lgrad2, circle, minimum width=0.2cm, minimum height=0.2cm, minimum size=0.2cm, draw=black, thick] (lai) {\scriptsize$\sim$};
		\node [right=0.02cm of lai, align=left] {\scriptsize Active\\[-0.9ex]\scriptsize Intercon-\\[-0.9ex]\scriptsize nection~~~~~~~~~~~~~~~~~~~~~¸};

	\end{tikzpicture}
	\vspace{-4mm}
	\caption{Gradients navigate through the network of components that encodes different regularities, dynamically modulated by changing information exchanges within the active interconnections. The colored paths represent relevant gradients for the experiment shown in  \Cref{fig:potentials}.}\label{fig:network}
\end{figure}
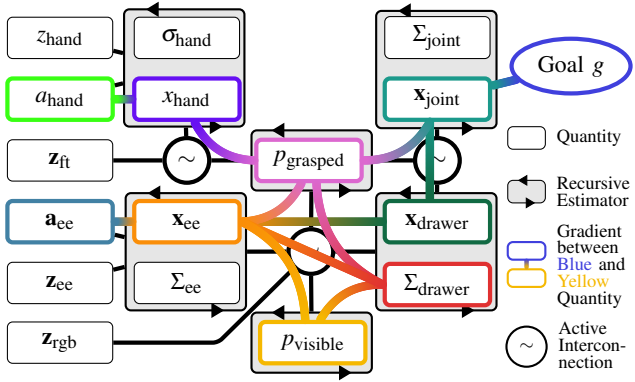

\begin{figure*}[t!]
	\includegraphics[width=\textwidth]{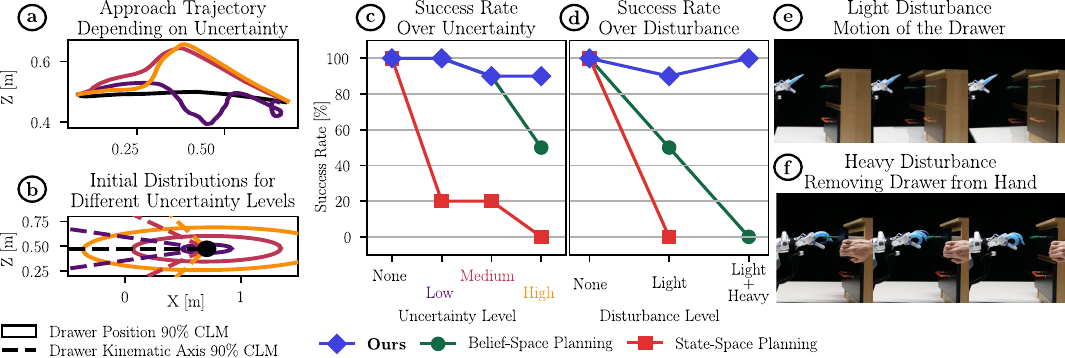}
	\vspace{-5mm}
	\caption{Our approach robustly resolves uncertainty and adapts to dynamic changes: If we vary the uncertainty under which the system operates (both by changing the initial prior as shown in (\textbf{b}) and adding sensor noise), our approach adapts its behavior to resolve uncertainty, e.g. by triangulating the drawer (\textbf{a}). This allows it to maintain performance under uncertainty outperforming planning approaches (\textbf{c}). If we instead introduce unexpected light (\textbf{e}) or heavy (\textbf{f}) disturbances, our approach can adapt to state changes and thus remains robust where planning approaches fail (\textbf{d}).}\label{fig:comparison}
	\vspace{-5mm}
\end{figure*}

\begin{figure}[b!]
	\includegraphics[width=\columnwidth]{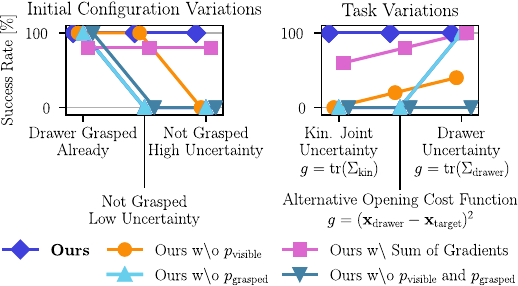}
	\vspace{-5mm}
	\caption{Active interconnections enable generalization over varying initial configurations and tasks: We evaluate our approach against versions using summed gradients and reduced active interconnections across three initial configurations and three alternative tasks. Summing gradients results in unstable, jerky behavior and occasional failures, while reducing active interconnections limits adaptability and generalization. Each system was tested $n=5$ times per setting.}\label{fig:ablations}
\end{figure}

\section{An \AICON-ic System for Drawer-Opening}\label{sec:drawer_exp}

In this section, we apply the \AICON-pattern to a partially observable drawer-opening task. By encoding necessary regularities in a network of actively interconnected estimators (\Cref{sec:drawer_network}), we derive gradients that represent different emerging subbehaviors that are automatically sequenced based on feedback from the environment (\Cref{sec:drawer_grads}). In \Cref{sec:drawer_quanti_experiment}, we demonstrate the advantages of our approach compared to planning approaches under uncertainty and in dynamic environments and then showcase the generality of our approach by studying different goals and initial configurations in \Cref{sec:drawer_experiemnt_generality}.

\subsection{Components and Active Interconnections}\label{sec:drawer_network}
For this task, we use a Panda robot with an RBO Hand 3~\cite{puhlmann_rbo_2022}, a wrist-mounted force-torque (FT) sensor, and an in-hand RGB camera. The robot’s proprioception provides estimates of the end-effector (EE) pose $\mathbf{x}_\mathrm{ee} \in \mathrm{se}(3)$ and hand state $x_\mathrm{hand} \in [0, 1]$, with $0$ representing a closed and $1$ an open hand. Both are tracked using extended Kalman filters (EKF) based on EE velocities $\mathbf{a}_\mathrm{ee}$ and hand inflation $a_\mathrm{hand}$.

To open a drawer, we need to estimate both the location of its handle $\mathbf{x}_\mathrm{drawer} \in \mathbb{R}^3$ and the parameters of its kinematic joint $\mathbf{x}_\mathrm{kin} = (\phi, \theta, q, \mathbf{p})$, where $\phi$ and $\theta$ are the joint axis orientation in spherical coordinates, $q$ is the translation along the axis, and  $\mathbf{p}\in \mathbb{R}^3$ is the closed position. The handle’s location is inferred from the camera image when visible or from the EE pose when grasped. Over time, we estimate the drawer's joint parameters from its motion. For both, we again use EKFs that also track uncertainty as covariance matrices $\boldsymbol{\Sigma}_\mathrm{drawer} \in \mathbb{R}^{3 \times 3}$ and $\boldsymbol{\Sigma}_\mathrm{kin} \in \mathbb{R}^{6 \times 6}$.

The estimator for the handle is in an active interconnection with the EE-pose estimator, since the interpretation of measurements from the wrist-mounted camera while visible and from its motion while grasped depend on the current EE pose. As this interaction drastically changes depending on whether the handle is currently visible, grasped, or neither, we estimate the scalar likelihoods $p_\mathrm{visible}$ and $p_\mathrm{grasped}$ and consider them in the two components' interconnection. To estimate $p_\mathrm{visible}$, we compare the angle of the handle's relative position to the camera's field of view. To estimate $p_\mathrm{grasped}$, we first estimate the current distance to a good grasping pose and if this distance is small also take into account the current hand state $x_\mathrm{hand}$ and FT measurements. 

The grasp likelihood $p_\mathrm{grasped}$ also regulates the active interconnection between the handle and kinematic joint estimators. When ungrasped, changes in handle motion update the closed location $\mathbf{p}$. When grasped, drawer movement reveals information about the joint axis, refining the estimates of $\phi$, $\theta$, and $q$, analogously to~\cite{martin-martin_coupled_2022}.

\subsection{Resulting Gradients and Behavior}\label{sec:drawer_grads}
Based on the described system of interconnected components, we can let the system open the drawer by specifying a goal and descending the steepest gradient. We encode opening a drawer as a quadratic cost over the distance of the current joint state $q$ to the desired one [$g(q)=(q-\SI{20}{\centi\meter})^2$] and thus obtain multiple gradient paths, shown in \Cref{fig:network} as colored paths.  They each compose some of the regularities encoded in the system to generate a useful subbehavior depending on the system's state. Since their magnitude is influenced by other estimated quantities through the active interconnections, we can select the steepest gradient to select the currently most appropriate one. 

As a result the robot's behavior follows a set of emerging behavioral rules: If the drawer is grasped, the robot can simply move it to the right location (\Cref{fig:potentials}e/g). If not, it has to first approach and grasp it (\Cref{fig:potentials}d). However, if the drawer location is uncertain, grasping will most likely fail, thus it needs to obtain more informative visual measurements by changing the viewpoint since the camera only obtains 2D images (\Cref{fig:potentials}a/c). This is only possible while the handle is visible. Otherwise, it can reestablish visibility (\Cref{fig:potentials}b/f). If state estimates change during execution either due to uncertainty or unexpected disturbances, the gradients and their magnitudes change and thus the behavior adapts (\Cref{fig:potentials}e-g). This lets our system perform well in uncertain and dynamic environments, which we will evaluate in comparison to planning in the following section.

\subsection{Experiments: Dynamically Composed Gradient Descent Resolves Uncertainty and Adapts to Dynamic Changes}\label{sec:drawer_quanti_experiment}
We evaluated our approach in real-world settings under varying uncertainty and dynamic conditions, running each scenario ten times. To simulate uncertainty, we varied the system’s prior knowledge of the drawer’s location and joint parameters  (\Cref{fig:comparison}b) and added artificial sensor noise. In high-uncertainty scenarios, the system adapted by triangulating the drawer’s location (\Cref{fig:comparison}a) and continuously adjusting its kinematic estimates, maintaining consistent performance across different levels of uncertainty (\Cref{fig:comparison}c). To test adaptability to dynamic changes, we introduced light disturbances (changing the cabinet’s pose, \Cref{fig:comparison}e) and heavy disturbances (removing the drawer from the hand post-grasp, \Cref{fig:comparison}f). In both cases, the system quickly adapted and maintained robust performance (\Cref{fig:comparison}d). Across these seventy trials, the system failed only three times, every time due to unmodeled issues like singularities and self-collisions.

We compared our method to two baseline planners: a state-space planner with a precomputed trajectory and a belief-space planner that refines estimates via exploratory motions (3 initial viewpoints and 2 kinematic joint explorations), similar to \cite{garrett_online_2020}. As shown in \Cref{fig:comparison}c-d, the belief-space planner handled uncertainty well but struggled with large disturbances, while the state-space planner performed poorly in both conditions. Our approach outperformed both, adapting more effectively while being computationally more efficient than 67-dimensional belief-space planning.

\subsection{Ablations and Variations: Active Interconnections Enable Generalization}\label{sec:drawer_experiemnt_generality}
Active interconnections in our system allow it to identify useful gradients depending on the world's state, enabling it to generalize across different configurations and tasks. We conducted ablation studies to compare the full system with variants that either reduced active interconnections or summed gradients instead of selecting only the steepest one.

As shown in \Cref{fig:ablations}, both the full system and the gradient sum variant handled the task variations, though the latter exhibited unstable, jerky behavior that occasionally led to failure. Disabling active interconnections results in sharp performance drops, particularly in tasks requiring view-point adjustment or grasping, as the system lost its adaptability.

\section{Discussion and Biological Similarities}\label{sec:discussion}

\paragraph*{Solvable Problem Classes and Limitations} 
Our results show that dynamically composed gradient descent robustly handles sequential tasks, even those with many steps. But as seen in Blocks World, the challenge lies in balancing competing subgoals and anticipating future opportunities, which may lead to suboptimal outcomes. Thus, tasks involving irreversible actions, e.g., destroying objects, or conflicting subgoals that require temporary setbacks, e.g., Towers of Hanoi, still require planning. Future research could explore adaptive methods for modifying subgoal hierarchies during execution, reducing the need for explicit planning.

\paragraph*{Biological Similarities in Behavior} 
The behavior emerging from our model mirrors human problem-solving, where real-time feedback often replaces explicit planning, even in tasks typically seen as planning problems. Humans take epistemic actions~\cite{kirsh_distinguishing_1994}, gathering information, unless forced into planning by task constraints~\cite{waldron_influence_2011}. Thus, suboptimalities appear in human tower-based tasks~\cite{waldron_influence_2011,mckinlay_planning_2008,kaller_reviewing_2011}, reflecting those in our approach. Interestingly, the limitations of our model align with those of Parkinson patients who struggle with goal hierarchies but not the amount of moves~\cite{mckinlay_planning_2008} and with the limited foresight observed in great apes solving sequential tasks~\cite{tecwyn_novel_2013}. In uncertain environments like the drawer task, our model, like humans~\cite{yeo_when_2016}, adapts dynamically to obtain more information, as seen in the emergence of triangulation.

\paragraph*{Biological Similarities in Structure}
Structurally, AICON resembles models of human information processing. As~\cite{martin-martin_coupled_2022} discusses, it shares similarities with human visual processing, which has led to its successful application in modeling human vision~\cite{battaje_information_2024,mengers_robotics-inspired_2024}. The behavior-generation process also parallels human gradient-based motor control, where there is even evidence for selecting the steepest gradient among multiple options~\cite{akulin_sloppy_2019}. The resulting adaptive potential field driven by component interactions is also a concept of biological interest~\cite{mcmillen_collective_2024}, while the integration of perception with action and goal selection aligns with theories on the prefrontal cortex~\cite{fine_whole_2022} and perception-action coupling in biological systems~\cite{bohg_interactive_2017}.

\section{Conclusion}

We introduced a gradient descent-based method for sequential tasks that dynamically composes regularities of the world to adjust both perception and action. This process considers subgoals implicitly, allowing it to handle multi-step problems, as long as the involved subgoals do not directly conflict. By leveraging feedback, it adapts to uncertainty and unexpected outcomes, leading to robust emergent behavior that leverages interactive perception and recovers from errors. This demonstrates the potential of computationally efficient, feedback-driven strategies in solving sequential tasks, aligning closely with observations on biological problem-solving behavior.

\newpage
\bibliographystyle{IEEEtran}
\bibliography{icra25}

\end{document}